\documentclass[]{biometrika}

\usepackage{amsmath}
\usepackage{smile}
\usepackage{hyperref} 
\usepackage{xr}

\usepackage{newtxtext}
\usepackage[subscriptcorrection]{newtxmath}
\usepackage{color}
\usepackage{mathtools}
\usepackage{cancel}
\usepackage{float}
\usepackage{booktabs} 
\usepackage{siunitx}  
\usepackage{multirow}

\graphicspath{{./art/}}

\usepackage[plain,noend]{algorithm2e}

\makeatletter
\renewcommand{\algocf@captiontext}[2]{#1\algocf@typo. \AlCapFnt{}#2} 
\def\@algocf@capt@plain{top}
\renewcommand{\algocf@makecaption}[2]{%
  \addtolength{\hsize}{\algomargin}%
  \sbox\@tempboxa{\algocf@captiontext{#1}{#2}}%
  \ifdim\wd\@tempboxa >\hsize
    \hskip .5\algomargin%
    \parbox[t]{\hsize}{\algocf@captiontext{#1}{#2}}
  \else%
    \global\@minipagefalse%
    \hbox to\hsize{\box\@tempboxa}
  \fi%
  \addtolength{\hsize}{-\algomargin}%
}
\makeatother

\def\reals{{\mathbb R}}

\def\G{{\mathcal G}}

\def\E{{\mathbb E}}
\def\argmin{\mathop{\text{\rm arg\,min}}}

\def\op{\text{op}}
\def\N{\text{N}}

\def\argmax{\mathop{\text{\rm arg\,max}}}

\def\span{{\mathop{\text{\rm span}}}}
\def\tr{{\mathop{\text{\rm tr}}}}
\def\vec{{\mathop{\text{\rm vec}}}}

\begin{document}



\markboth{Transfer Matrix Completion}{Transfer Matrix Completion}

\title{Representational Transfer Learning for  Matrix Completion}

\author{Yong He}
\affil{Institute for Financial Studies, Shandong University, China.
\email{heyong@sdu.edu.cn}}

\author{Zeyu Li}
\affil{School of Management, Fudan University, China.
\email{zeyuli21@m.fudan.edu.cn}}

\author{Dong Liu}
\affil{  School of Physical \& Mathematical Sciences, \\Nanyang Technological University, Singapore.}

\author{Kangxiang Qin}
\affil{Institute for Financial Studies, Shandong University, China.}

\author{\and Jiahui Xie}
\affil{Department of Statistics and Data Science,\\ National University of Singapore,  Singapore.}

\maketitle

\begin{abstract}
We propose to transfer representational knowledge from multiple sources to a target noisy matrix completion task by aggregating singular subspaces information. Under our representational similarity framework, we first integrate linear representation information by solving a two-way principal component analysis problem based on a properly debiased matrix-valued dataset. After acquiring better column and row representation estimators from the sources, the original high-dimensional target matrix completion problem is then transformed into a low-dimensional linear regression, of which the statistical efficiency is guaranteed. A variety of extensional arguments, including post-transfer statistical inference and robustness against negative transfer, are also discussed alongside. Finally, extensive simulation results and a number of real data cases are reported to support our claims.
\end{abstract}


\section{Introduction} The basic idea of transfer leaning, also known as knowledge transfer, is to leverage information from related sources to improve the statistical performance of the target study of interest \citep{torrey2010transfer,zhuang2020comprehensive}. In high-dimensional statistical settings, a flourish of transfer learning methods have been proposed with various types of similarity assumptions. For example, when the differences between the target and source parameters, also known as contrast vectors, are sufficiently sparse, \cite{bastani2021predicting, li2020transfer,tian2022transfer,transquantile} propose to use a two-step procedure, i.e., pooling and debiasing, to better estimate the target parameter. \cite{li2023estimation} further suggest joint estimation of the target parameters and the contrast vectors, so as to relax the homogeneous Hessian matrices condition, see also \cite{li2024simultaneous} for a joint estimation method by a non-convex penalty. Please refer to \cite{cai2021transfer,reeve2021adaptive,yan2024transfer} for transfer learning in other statistical problems including non-parametric classification and causal inference.

In this work, the target task of interest is a matrix completion problem contaminated by noises. To be more specific, let $\Theta_0^*\in \reals^{p\times q}$ be the unknown target matrix, for $i\in[n_0]:=\{1,\dots,n_0\}$, we have i.i.d. observations of the form 
\begin{equation}\label{eq:matcomp1}
    y_{0,i} = \left(\Theta_0^*\right)_{a_{0,i},b_{0,i}} + (pq)^{-1/2}\xi_{0,i},
\end{equation}
where $\xi_{0,i}$ has zero mean and variance $\sigma_0^2$, and $(a_{0,i},b_{0,i})$ are randomly chosen indices. The rescaling factor $(pq)^{-1/2}$ originates from \cite{negahban2012restricted}, to ensure a constant signal-to-noise ratio when $\Theta_0^*$ has a constant Frobenius norm regardless of the matrix dimensions. Alternatively, we have an equivalent formulation of (\ref{eq:matcomp1}) by defining $X_{0,i}=e_{a_{0,i}}e_{b_{0,i}}^{\top}$ where both $e_{a_{0,i}}$ and $e_{b_{0,i}}$ are taken from the natural basis of $\reals^p$ and $\reals^q$, respectively. The observation model then becomes
\begin{equation}\label{eq:matcomp2}
   y_{0,i} = \left\langle \Theta_0^*, X_{0,i}\right\rangle +(pq)^{-1/2}\xi_{0,i}, \quad i\in[n_0].
\end{equation}

The trace regression formulation (\ref{eq:matcomp2}) is insightful enough to offer a way to estimate $\Theta_0^*$, that is, by minimizing the mean square error. However, it is statistically impossible to retrieve the $p \times q$ parameters using merely $n_0$ observations when $n_0\ll p q$, unless additional low-rank assumptions are made on $\Theta_0^*$.  \cite{candes2012exact} relax the low-rank constraint of a matrix using the nuclear norm of the matrix $\|\cdot\|_{\N}$, which has also spawned a variety of penalized least squares methods of the nuclear norm \citep{candes2010matrix,candes2010power,negahban2011estimation,negahban2012restricted,cai2015rop,cai2016matrix}. These computationally tractable convex relaxed methods are known to be statistically optimal under mild conditions. In parallel, another stream of work focuses on non-convex approaches including matrix factorization. For example, \cite{keshavan2009matrix} propose a method based on singular value decomposition (SVD); see also \cite{zhao2015nonconvex,wang2017unified,ma2018implicit,chen2020noisy}.

In the context of knowledge transfer, we have $K$ additional source studies with underlying true matrices $\Theta_k^*\in\reals^{p\times q}$ for $k\in [K]$. In some cases, these $\Theta_k^*$ are fully observed, probably only contaminated by  random noises $E_k$. That is, we can directly observe $\tilde{\Theta}_k=\Theta_k^*+E_k$. However, in the main body of this work we consider a more challenging setting where these $\Theta_k^*$ are only partially observed as well, leaving the fully observed cases to the extensional arguments in the supplementary material. Specifically, we set $X_{k,i}=e_{a_{k,i}}e_{b_{k,i}}^{\top}$, and 
$$ y_{k,i} = \left\langle \Theta_k^*, X_{k,i}\right\rangle +(pq)^{-1/2} \xi_{k,i},\quad i\in[n_k],$$ where $\xi_{k,i}$ has zero mean and variance $\sigma_k^2$. Following \cite{xia2021statistical}, we independently and uniformly sample the random matrices $X_{k,i}$ from the canonical orthogonal basis $\mathcal{E}:=\{e_{a} e_b^{\top}: a \in[p], b \in[q]\}$ for $k\in\{0\}\cup [K]$. We shall always assume that $p\geq q$ without loss of generality.

\subsection{Similarity metrics}

The characterization of the similarity metric between the target and source studies is essential before tailoring any knowledge transfer algorithms. The similarity metric pins down the shareable features between these statistical tasks, and we design the corresponding algorithm to fully exploit the shared common information, so as to enhance the statistical performance of the target studies. For the sake of brevity in presentation, we shall first assume that all sources are informative, i.e., close to the target, under certain sense of similarity.

A commonly used and easy-to-analyze form of similarity is the distance similarity. In the context of matrix completion, distance similarity shall require $\|\Theta_0^*-\Theta_k^*\|_{\N}\leq h_d$ for $k\in[K]$, where $\|\cdot\|_{\N}$ is the matrix nuclear norm and characterizes the low-rankness of contrast matrices $\Theta_0^*-\Theta_k^*$. Distance similarity often leads to the classical pooling and debiasing methods, where in the first step all source samples are gathered for a joint estimation, and in the second step the pooling estimator is debiased using only the target samples \citep{li2020transfer,tian2022transfer,transquantile}. However, in many practical cases, the distance similarity is clearly too ideal to be realistic, as we in essence assume the parameter matrices to be almost element-wise alike with the difference matrix being quite sparse or low-rank.

In the following, we give a few conceptual counter-examples of the distance similarity in empirical applications, where a more suitable similarity metric ought to be considered. First, most knowledge transfer methods in the literature require the dimension of the target parameter to be the same as the source parameters. In many real-world applications, however, we might encounter dimension mismatch due to the difference of measurement or storage  between target and the source datasets. That is, the shape of the target matrix $\Theta_0^*$ might not be the same as the source matrices $\Theta_k^*$ for $k\in[K]$. For example, in the COVID-19 computed tomography (CT) example used in this work, the sizes of chest scan images range from $p\in [115,1671]$ and $q\in [61,1225]$, in terms of the number of pixels in rows and columns. Though we may reshape these the matrices into the same size via upsampling, subsampling, or cropping, the distance similarity could be quite sensitive to these operations. Second, in some other cases, even if the target and source matrices are naturally the same size, the similarity would be more reasonable and inherent in the spectral sense, rather than in the distance way, for example the applications on multiple networks and multiple-input multiple-output (MIMO) radars \citep{zheng2022limit,sun2015mimo}.

\subsection{Representational similarity}

In this work, we adopt the representational similarity (or spectral similarity). First, write the singular value decomposition (SVD) of the rank $r_0$ matrix $\Theta_0^* = U^*_0\Lambda^*_0(V^*_0)^{\top}$, where $(U^*_0)^{\top}U^*_0=(V^*_0)^{\top}V^*_0=I_{r_0}$ and $\Lambda^*_0$ is diagonal. Although $\Theta_k^*$ might be far away from $\Theta_0^*$ in the distance sense such that $\|\Theta_k^*-\Theta_0^*\|_{\N}\gg h_d$, it could still contain valuable information on $U^*_0$ and $V^*_0$. To better illustrate our motivation, now suppose we are able to exactly recover $U^*_0$ and $V^*_0$ with the help of the auxiliary sources, the task of estimating $\Theta_0^*$ then changes
\begin{equation}\label{eq:high2low}
    \text{from}\quad \argmin_{\Theta_0\in \reals^{p\times q},\, \text{rank}(\Theta_0)=r_0}\E\left[y_{0,i}-\tr\left(X_{0,i}^{\top}\Theta_0 \right)\right]^2\quad \text{to}\quad \argmin_{\Lambda_0\in \reals^{r_0\times r_0}} \E\left[y_{0,i}-\tr\left(Z_{0,i}^{\top}\Lambda_0 \right)\right]^2,
\end{equation}
where $Z_{0,i}:=(U^*_0)^{\top}X_{0,i}V^*_0$ of size $r_0\times r_0$ is the ``extracted feature matrix" after acquiring knowledge about the low-dimensional linear representations $U^*_0$ and $V^*_0$. The original high-dimensional trace regression problem of dimension $p\times q$ is thus turned into a low-dimensional linear regression problem with at most $r_0\times r_0$ parameters, whose statistical efficiency is guaranteed by classical linear regression theory, avoiding the curse of dimensionality when $p$ and $q \rightarrow \infty$. To some extent, the debiasing step under distance similarity solves a relatively easier high-dimensional statistical task, so the gain of knowledge transfer for the classical two-step methods is from high-dimensions to high-dimensions \citep{li2020transfer,tian2022transfer,transquantile}. On the contrary, the gain via representational transfer could be from high-dimensions to low-dimensions, which seems to be more promising in principle. 

We now present our knowledge transfer framework. For brevity, we shall assume first that all source datasets are informative, and we leave the extensions against potentially harmful sources to Section \ref{sec:nora}. Now for $\cI:=[K]$ and $k\in \{0\}\cup\cI$, let $\Theta_k^*$ be of rank $r_k$ with the singular value decomposition:
\begin{equation}\label{equ:thetak-svd}
  \underbrace{\Theta_k^*}_{p\times q}=\underbrace{U^*_k}_{p\times r_k}\times \underbrace{\Lambda^*_k}_{r_k\times r_k}\times \underbrace{\left(V_k^*\right)^\top}_{r_k\times q},
\end{equation}
where $(U^*_k)^{\top}U^*_k=(V^*_k)^{\top}V^*_k=I_{r_k}$ and $\Lambda^*_k$ is diagonal. We allow $r_k$ to differ across different studies. Before we characterize the similarity between the linear representations $U_k^*$ and $V_k^*$, note that (\ref{equ:thetak-svd}) naturally inherits the identifiability problem such that $U_0^*$ could be replaced by $U_0^*H_k$ for any $r_k\times r_k$ orthogonal matrix $H_k$ while changing $\Lambda_k^*$ to $H_k^{\top}\Lambda_k^*$ correspondingly. However, the subspaces $\span(U_k^*)$ and $\span(V_k^*)$ are still identifiable and unique, so it is more natural to impose the representation similarity in the subspace sense.

Due to symmetry, we first focus on $\span(U_k^*)$. Although $\span(U_k^*)$ might in general be quite different from $\span(U_0^*)$ for $k\in \cI$, we assume that $\span(U^*_0)$ lies in a larger subspace of dimensions $p_0$ of $\reals^{p}$, denoted as $\span(U)$, which also approximately contains all $\span(U_k^*)$ for $k\in \cI$, so that $p_0\geq \max_{k\in\cI}(r_k)$. To mathematically quantify such ``approximately contains", let $U_{\perp}$ be the orthogonal complement of $U$, for some small $h>0$, we assume that
\begin{equation}\label{eq:rep-sim1}
  \span(U_0^*)\subseteq\span(U),\quad  \left\|U_{\perp}^{\top}U^*_k\right\|_F \leq h, \quad k\in \cI.
\end{equation}
Geometrically, the principal angles between $\span(U_{\perp})$ and $\span(U^*_k)$ are $\{\cos^{-1}(\sigma_{k,l})\}_{l=1}^{r_k}$ where $\sigma_{k,l}$ are non-trivial singular values of $U_{\perp}^{\top}U^*_k$. Hence, small $\|U_{\perp}^{\top}U^*_k\|_F$ means that $\span(U^*_k)$ and $\span(U_{\perp})$ are almost orthogonal, so that $\span(U^*_k)$ is almost within $\span(U)$. Similar arguments could be made with respect to $\span(V^*_k)$, where we assume that there exists some $q_0$ dimensional $\span(V)$ with $q_0\geq \max_{k\in\cI}(r_k)$, such that for its orthogonal complement $V_{\perp}$ we assume

\begin{equation}\label{eq:rep-sim2}
\span(V_0^*)\subseteq\span(V),\quad \left\|V_{\perp}^{\top}V^*_k\right\|_F \leq h, \quad k\in \cI.
\end{equation}

Clearly, $\span(U)=\span(V)=\reals^p$ is a trivial case with $h=0$, where, however, we cannot achieve dimension reduction and hence achieve no gain of knowledge transfer as depicted by (\ref{eq:high2low}). It is foreseeable that $h$ gets larger when we seek smaller $p_0$ and $q_0$, that is, a stronger dimension reduction. Basically, when $p_0$ and $q_0$ become larger, the estimated linear representations are more capable of capturing the target matrix structure of interest with a smaller bias $h$. However, enlarging $p_0$ and $q_0$ at the same time increases the statistical difficulty in estimating the post-representation target coefficients, since the number of parameters to be estimated by this linear regression is $p_0q_0$. Therefore, there is essentially a trade-off when choosing the cut-off dimensions $p_0$ and $q_0$ here. Throughout this work, we shall assume a sufficiently small $h$ given finite $\max(p_0,q_0)<\infty$. Similar linear representational conditions have also been assumed in \cite{tian2023learning}, where the authors add a subspace penalty to enforce similar representations when learning all tasks by aggregating the data. Meanwhile, in \cite{he2024representation}, a non-linear representation function is assumed to be shared across various tasks and is estimated in a joint way. 

\subsection{Contributions and notations}

The contributions of this work lie in the following aspects. First, as an addition to the high-dimensional knowledge transfer literature, by introducing two-way representational similarity into the matrix completion problem, we manage to decompose the original high-dimensional statistical task into two parts. In the first part, we carefully extract the useful source information via unsupervised subspace integration methods. After the representations are recovered, the second part for the target dataset is inherently low-dimensional and statistically efficient. It is of independent importance for the high-dimensional transfer learning community, as for most two-step methods, the debiasing step often involves a relatively easier but still high-dimensional statistical task, for example, estimating a even sparser high-dimensional contrast vector \citep{li2020transfer,tian2022transfer,transquantile}. Detailed theoretical discussions and extensive numerical results support our claims above. Second, as one can always stack the target and $K$ source matrices into a tensor of size $(K+1)\times p\times q$, this work can also be viewed as an extension to the well-developed field of low-rank tensor completion; see, for example, \cite{yuan2016tensor,yuan2017incoherent,xia2021statistically}. In contrast to the existing tensor completion methods, we assume only the low-rankness in the latter two modes. Meanwhile, after acquiring the representations, our target estimation step shows a great computational advantage. It is particularly useful for streaming data, where the target samples arrive sequentially, as it makes little sense to complete a large tensor immediately each time a new matrix arrives. In the end, we carefully tailor our algorithm to avoid the inclusion of potentially harmful source datasets. As validated by our experiments, it might cause severe negative transfer if one blindly applies tensor completion methods to the stacked tensor.

To end this section, we introduce some notations used throughout the paper. For a real matrix $A$, let $\{\lambda_i(A)\}$ be its non-increasing singular values and $d_i(A) = \lambda_i(A)-\lambda_{i+1}(A)$ be the $i$-th singular value gap. The operator norm of the matrix $A$ is denoted by $\|A\|_{\op}=\lambda_1(A)$, while its Frobenius norm is written as $\|A\|_{F}$. For a random variable $X\in \reals$, we define $\|X\|_{\psi_2}=\sup_{p\geq 1}p^{-1/2}(\E|X|^p)^{1/p}$ and $\|X\|_{\psi_1}=\sup_{p\geq 1}p^{-1}(\E|X|^p)^{1/p}$. Please, see \cite{vershynin2018high} for details of the sub-Gaussian and sub-exponential norms. Moreover, $o_{p}$ in this work is for convergence to zero in probability and $O_{p}$ means stochastic boundedness. We write $x\lesssim y$ if $x\leq Cy$ for some $C>0$, $x\gtrsim y$ if $x\geq cy$ for some $c>0$, and $x\asymp y$ if both $x\lesssim y$ and $x\gtrsim y$ hold. The constants may not be identical across different lines.

\section{Oracle Transfer Learning Procedure}
We present an oracle three-step procedure designed specifically for the representation similarity between the target and source matrix completion studies, assuming that  all available sources are informative. Generally speaking, in the first step, we acquire debiased estimators of $\Theta_k^*$ for $k\in \cI:=[K]$, where $\cI$ represents informative sources. We denote the debiased estimators by $\tilde{\Theta}_k$ and eventually construct an unbiased matrix-valued dataset $\cD_{\cI} :=\{\tilde{\Theta}_k, k\in \cI\}$. As these $\tilde{\Theta}_k$ shall share linear representation information within our framework, retrieving subspace population information from $\cD_{\cI}$ then turns into a standard two-way principal component analysis problem \citep{zhang20052d,wang2019factor,chen2021statistical}. That is to say, in the second step we could take an unsupervised learning approach to integrate the subspace information across the unbiased $\tilde{\Theta}_k$, to acquire the $p\times p_0$ column orthogonal matrix $\hat{U}$ and the $q\times q_0$ column orthogonal matrix $\hat{V}$, which are also called linear representations. In the third step, we turn the original high-dimensional target low-rank trace regression problem to a low-dimensional linear regression with the better estimated linear representations as alluded in (\ref{eq:high2low}), to achieve an efficiency gain in the target performance. To be more specific, after acquiring $\hat{U}$ and $\hat{V}$, we go back to the target task and calculate the projected extracted feature matrices $\hat{Z}_{0,i}=\hat{U}^{\top}X_{0,i}\hat{V}$ of shape $p_0\times q_0$. We perform a linear regression on the projected dataset $\{(\hat{Z}_{0,i},y_{0,i})\}_{i=1}^{n_0}$, to acquire $\hat{\Gamma}_0$. In the end, we retrieve our target matrix completion estimator as $\hat{\Theta}_0 = \hat{U}\hat{\Gamma}_0\hat{V}^{\top}$.

    


In the main body of this work, source information is treated as externally given for the target task, which is a common practice for transfer learning. Hence, we do not include a debiased $\tilde{\Theta}_0$ in the first step, but this is totally feasible. Meanwhile, the construction details of the unbiased dataset for partially observed sources in the first step and the choice of the two-way principal component analysis method in the second step are still unclear. Though in principle these two steps are totally flexible and could be tailored according to each specific problem at hand, we present our default implementation procedure as follows.

\subsection{Default implementation procedure}\label{sec:ori}

First, the unbiasedness of $\tilde{\Theta}_k$ in the first step is essential to establish the theoretical properties of any integrated subspace estimator in the second step. When the sources are partially observed, most existing matrix completion estimators, if not all, are inherently biased \citep{candes2010matrix,cai2015rop,wang2017unified,chen2020noisy}. Hence, it is important to first debias any crude estimators we acquire using these existing methods. The debiasing technique is important for the statistical inference of sparse linear regression problems \citep{javanmard2014confidence,van2014asymptotically}. We may also refer to \cite{carpentier2018iterative,carpentier2018adaptive,xia2019confidence} for low-rank matrix trace regression and to \cite{chen2019inference,xia2021statistical} for  noisy matrix completion problems. 

We adopt the estimation and then debiasing procedure from \cite{xia2021statistical} via $J$-fold sample splitting. For notational brevity, without loss of generality, we assume that $n_k/J$ is an integer for all $k\in \cI$. Then we split the indices of the $k$-th dataset $D_k:=[n_k]$ into $J$ non-overlapping sub-indices $D^j_{k}$ for $j\in J$, each of length $n_k/J$. We also write $D^{-j}_{k}$ as the sub-indices of $D_k$ without $D^j_{k}$, i.e., $D^{-j}_{k}=D_k\setminus D^j_{k}$. We denote $D^j_{k}$ and $D^{-j}_{k}$ as the corresponding sub-samples of the $k$-th dataset without causing confusions. Then, for each $j\in J$, we first acquire a crude matrix completion estimator $\dot{\Theta}^{-j}_k$ by a certain reliable method (generally biased) using $D_k^{-j}$. Then, we debias $\dot{\Theta}^{-j}_k$ using $D^j_{k}$ according to
\begin{equation}\label{eq:debias}
  \tilde{\Theta}_k^j:=\dot{\Theta}^{-j}_k+\frac{Jpq}{n_k}\sum_{i\in D_k^j}\left(y_{k,i}-\left\langle \dot{\Theta}^{-j}_k,X_{k,i}\right\rangle\right)X_{k,i}.
\end{equation}
Finally, we let $\tilde{\Theta}_k:=\sum_{j\in[J]}\tilde{\Theta}_k^j/J$.

That is, by the debiasing procedure, we now have an unbiased matrix-valued dataset $\cD_{\cI}=\{\tilde{\Theta}_k\}_{k\in \cI}$ that includes $K$ matrices. Then it becomes a two-way principal component analysis problem that aims to retrieve the population column and row subspaces of the matrices in $\cD_{\cI}$. For this problem, we also have a rich toolbox of methods, for example, those introduced in \citep{zhang20052d,wang2019factor,chen2021statistical, Yu2021Projected,2dgb}. As a simple illustration, by the benchmark method $(2D)^2$-principal component analysis (PCA) in \cite{zhang20052d}, we can estimate $\span(U)$ by extracting the leading $p_0$ eigenvectors of the matrix $\sum_{k\in\cI} n_k\tilde{\Theta}^{j}_k(\tilde{\Theta}^{j}_k)^{\top}/N$, where $N=\sum_{k\in \cI}n_k$. In this paper, we suggest using the one-round Grassmannian barycenter method from \cite{2dgb}, due to its robustness against outliers and communication efficiency. For each matrix-valued data $\tilde{\Theta}_k$, we first take its leading $r_k$ left and right singular vectors, that is, those corresponding to the largest $r_k$ singular values, and stack them into $p\times r_k$ and $q\times r_k$ column orthogonal matrices $\tilde{U}_k$ and $\tilde{V}_k$. We then calculate the orthogonal projection matrices corresponding to the subspaces spanned by these eigenvectors as $\tilde{P}_k:=\tilde{U}_k(\tilde{U}_k)^{\top}$ and $\tilde{Q}_k:=\tilde{V}_k(\tilde{V}_k)^{\top}$. In the end, we integrate the subspace information by extracting the leading $p_0$ and $q_0$ eigenvectors of the weighted Euclidean mean of the projection matrices
\begin{equation}\label{eq:GB-os}
    \tilde{\Sigma}_U:=\frac{1}{N} \sum_{k\in\cI} n_k \tilde{P}_k, \quad \tilde{\Sigma}_V:=\frac{1}{N} \sum_{k\in\cI} n_k \tilde{Q}_k,
\end{equation}
denoted as $\hat{U}$ and $\hat{V}$ respectively. 

\begin{algorithm}
\caption{Default implementation procedure under partially observed sources.}\label{alg:one-round}
\begin{algorithmic}[1]
\REQUIRE ~~\\
    $(X_{k,i}, y_{k,i})$ for $k\in\cI= [K]$; cut-off dimensions $\{r_k\}_{k\in\cI}$ and $p_0$, $q_0$; number of folds $J$;\\
\ENSURE ~~\\
    \STATE 1. debiased dataset: for each $k\in\cI$, first split the dataset $D_k$ into $J$ folds. For each $j\in [J]$, first acquire a crude estimator $\dot{\Theta}_k^{-j}$ using $D_k^{-j}$, debias $\dot{\Theta}_k^{-j}$ using $D_k^{j}$ according to (\ref{eq:debias}), and then acquire the unbiased matrix $\tilde{\Theta}_k=\sum_{j\in[J]}\tilde{\Theta}^{-j}_k/J$, so as to construct the matrix-valued dataset $\cD_{\cI} =\{\tilde{\Theta}_k, k\in \cI\}$;\\
    
    \STATE 2. subspace integration: acquire $\hat{U}$ and $\hat{V}$ by taking the leading $p_0$ and $q_0$ eigenvectors of $\tilde{\Sigma}_U$ and $\tilde{\Sigma}_V$ given in (\ref{eq:GB-os});
    \STATE 3. target estimation: apply linear regression on $\{(\hat{Z}_{0,i},y_{0,i})\}$ with $\hat{Z}_{0,i}=\hat{U}^{\top}X_{0,i}\hat{V}$ to obtain the $p_0\times q_0$ parameter matrix $\hat{\Gamma}_0$;\\
    
\RETURN $\hat{\Theta}_{0}= \hat{U}\hat{\Gamma}_0\hat{V}^{\top}$.
\end{algorithmic}
\end{algorithm}

The default implementation procedure under partially observed sources is summarized in Algorithm \ref{alg:one-round}. Apparently, the performance of our knowledge transfer method depends on the statistical precision of the crude estimators $\{\dot{\Theta}_k^{-j}\}^{j\in[J]}_{k\in\cI}$. Fortunately, as pointed out in \cite{xia2021statistical} and shown in the following section, such a dependence is fairly weak as long as the estimation errors of the crude estimators are small in terms of the vectorized $\ell_\infty$ norm, which is denoted by $\|\cdot\|_{\infty}$ in this article. Recall that $\sigma_k^2$ is the variance of $\xi_{k,i}$, we adopt the following assumption directly from Assumption 1 of \cite{xia2021statistical}.

\begin{assumption}\label{assum:1}
    For $j\in [J]$ and $k\in [K]$, as $n_k, p,q\rightarrow \infty$, the crude estimators satisfy
    \begin{equation}\label{eq:assum1}
        \left\|\dot{\Theta}_k^{-j}-\Theta_k^*\right\|_{\infty}=o_p\left[ (pq)^{-1/2}\sigma_k\right].
    \end{equation}
\end{assumption}

Estimation error bounds similar to (\ref{eq:assum1}) can also be found in \cite{ma2018implicit,chen2019inference}. For concreteness, we resort to the non-sample-splitting version of a rotation calibrated Grassmannian gradient descent algorithm given by \cite{xia2021statistical}. It serves as our default choice for the matrix completion method in this work. A sample-splitting version of this algorithm is capable of producing crude estimators that satisfies Assumption \ref{assum:1} under some mild incoherence and noise conditions in \cite{xia2021statistical}. Meanwhile, as shown by extensive numerical experiments, the default implementation procedures stated in Algorithm \ref{alg:one-round} perform quite satisfactorily.

The debiasing strategy in Algorithm \ref{alg:one-round} originates from \cite{chernozhukov2018double} and is used in \cite{xia2021statistical} with $J=2$. This particular form of debiasing takes advantage of the fact that $\E[\vec(X_{k,i})\vec(X_{k,i})^{\top}]=(pq)^{-1}I_{pq}$, and shall be calibrated accordingly if the entries are not sampled uniformly. The $J$-fold sample-splitting technique avoids the loss of statistical efficiency associated with the sample splitting. Throughout this work, we consider $J$ as a finite integer. In the end, the choice of the debiasing method is clearly flexible as well; see also \cite{chen2019inference} for a similar debiased estimator under uniform sampling observations.

Finally, we briefly remark on the advantages of the selected one-round Grassmannian barycenter approach for the subspace integration step. The first advantage is its robustness against individual outliers that have extremely large eigenvalues, as the Grassmannian barycenter discards all eigenvalue information and only integrates the directional subspace information \citep{2dgb}. Robustness is extremely important in the current knowledge transfer context, so as to prevent negative transfer when accidentally including non-informative sources. Second, when multiple source datasets are distributed across different machines, the Grassmannian barycenter is not only communication-efficient but also privacy-protected as it only requires transmitting the local subspace estimators, namely $\tilde{U}_k^j$ and $\tilde{V}_k^j$, once, rather than transmitting the column and row covariance matrices or even the whole datasets. This advantage is also discussed in detail for distributed principal component analysis by \cite{fan2019distributed}.

\section{Oracle Theory}
In this section, we provide theoretical justification for the default implementation in Algorithm \ref{alg:one-round}. We first make some technical assumptions.

\begin{assumption}\label{assum:2}
    For $k\in \{0\}\cup[K]$, the noise $\xi_{k,i}$ is a centered sub-Gaussian random variable independent from the uniformly sampling $X_{k,i}$, satisfying $\|\xi_{k,i}\|_{\psi_2}\lesssim \sigma_k \leq C$.
\end{assumption}

\begin{assumption}\label{assum:3}
    For $k\in \{0\}\cup[K]$, $\Theta_k^*$ is of rank $r_k\leq r\leq \min(p_0,q_0)\leq\max(p_0,q_0) <\infty$, with non-trivial singular values $0<c\leq \lambda_{r_k}(\Theta_k^*)\leq\dots\leq \lambda_{1}(\Theta_k^*) \leq C$.
\end{assumption}

\begin{assumption}\label{assum:4}
    Assume that there exists some sufficiently small $h\geq 0$ such that both (\ref{eq:rep-sim1}) and (\ref{eq:rep-sim2}) hold for $k\in \cI$. Meanwhile, for $P_k^*=U^*_k(U^*_k)^{\top}$ and $Q_k^*=V^*_k(V^*_k)^{\top}$, assume there exists some $g>h$ that for any $u\in\span(U)$ and $v\in\span(V)$,
    $$\frac{1}{N}\sum_{k\in\cI} n_k\left\langle u, P_k^*u\right\rangle\geq g^2,\quad \frac{1}{N}\sum_{k\in \cI} n_k\left\langle v, Q_k^*v\right\rangle\geq g^2.$$
\end{assumption}

Assumptions \ref{assum:2} and \ref{assum:3} ensure a well-conditioned noise and signal, which are standard in the literature \citep{xia2021statistical}. Meanwhile, Assumption \ref{assum:4} is related to the identifiability of the population subspaces $\span(U)$ and $\span(V)$ using the Grassmannian barycenter method. If $\sum_{k=0}^{K} n_k\langle u, P_k^*u\rangle/N=0$ for some $u\in \reals^p$, it essentially means that $u$ is orthogonal to all $\span(U_k^*)$ and shall be identified as a vector in $\span(U_\perp)$ rather than $\span(U)$. The same argument applies to $\span(V)$. One may refer to \cite{2dgb} for detailed discussions on a similar identifiability condition. We present our main theoretical claim as follows. 

\begin{theorem}[Representation transfer learning]\label{theorem:main}
    Under Assumptions \ref{assum:1} to \ref{assum:4}, for $k\in \cI$, assume that $n_k\gtrsim p\log^{\tau} p$ for some $\tau\ge3$ and $K\lesssim (N/p\log p)^{1/2}$ as $n_0, n_k, p,q\rightarrow \infty$, we have
    \begin{equation}\label{eq:main-rate}
        \left\|\hat{\Theta}_0-\Theta_0^*\right\|^2_F=O_p\bigg(\underbrace{\frac{rp\log p}{N}+h^2}_{\text{representation}}+\underbrace{\frac{p_0q_0}{n_0}}_{\text{target}}\bigg).
    \end{equation}
\end{theorem}

We end this section with a few remarks. First, a typical rate for matrix completion methods that use only the target dataset should be $rp\log p/n_0$, while the information theoretical lower bound is of order $rp/n_0$ \citep{negahban2012restricted}. The first term of (\ref{eq:main-rate}) essentially matches the nearly information theoretical optimal rate with the dimensional factor $\log p$ given the pooled sample size $N=\sum_{k\in\cI}n_k$, which means that integrating the representation information alone is adequate for efficient and almost optimal knowledge transfer. The second term $h^2$ is standard and vanishes if all $\span(U_k^*)\subseteq \span(U)$ and $\span(V_k^*)\subseteq \span(V)$. Finally, the third term comes from the low-dimensional ordinary least squares after acquiring the low-dimensional representations using the target dataset. As alluded in the Introduction, as $p_0$, $q_0$ grows, we shall have smaller $h$ in principle. For example, if $p_0=p$ and $q_0=q$, then $h=0$ naturally. Representation learning thus requires a trade-off between $h^2$ in the representation error and the target estimation error $p_0q_0/n_0$. We conclude that representational transfer learning outperforms target estimation if $p_0q_0/n_0+h^2\ll rp\log p/n_0$ and $N\gg n_0$, which is mild as relatively small $p_0$ and $q_0$ would guarantee small $h$ in practice. In the end, we additionally require that $K\lesssim (N/p\log p)^{1/2}$. This is due to the fact that the Grassmannian barycenter method is inherently biased up to a high-order term. We restrict the size of the useful sources $K$ so that the high-order bias term could be absorbed into the variance term $p\log p/N$ for notational brevity. It still allows $K$ to diverge as long as $N\gg p\log p$, which is almost always the case. 

After acquiring the convergence rate of $\hat{\Theta}_0-\Theta_0^*$ in Theorem \ref{theorem:main}, we now discuss post-transfer inference given sufficiently reliable sources. We are interested in the asymptotic distribution of the bilinear form $\langle u,(\hat{\Theta}_0-\Theta_0^*) v\rangle$. A motivating example would be $u=e_i$ and $v=e_j$, which is particularly useful since we can perform statistical inference and construct a confidence interval for the $(i,j)$-th element of $\Theta_0^*$. We claim the following corollary of Theorem \ref{theorem:main}, while emphasizeing that the left hand side of \eqref{eq:tmc-an} as a whole could be directly calculated with observations.

\begin{corollary}[Asymptotic normality]\label{theorem: asymptotic normality}
Under the assumptions of Theorem \ref{theorem:main}, for some constant $\sigma_0>0$, if we further assume that $\sigma_0^2\min(\|U^{\top}u\|^2_2,\|V^{\top}v\|^2_2)/n_0\gg rp\log p/N +h^2 $ as $n_0, n_k, p,q\rightarrow \infty$, then
\begin{equation}\label{eq:tmc-an}
    \frac{\sqrt{n_0}}{\hat{\sigma}_L}\left\langle u,(\hat{\Theta}_0-\Theta_0^*)v\right\rangle \rightarrow \mathrm{N}(0,1)\; \text{in distribution,}\quad \text{where}
\end{equation}
$$\hat{\sigma}_L^2=\frac{pq}{n_0} \sum_{i=1}^{n_0}\left[y_{0,i}-\tr\left(X_{0,i}^{\top}\hat{\Theta}_0 \right)\right]^2\left\|\hat{U}^{\top}u\right\|_2^2\left\|\hat{V}^{\top}v\right\|_2^2.$$
\end{corollary}

\section{Non-Oracle Transfer Learning Procedure}\label{sec:nora}

 The preceding discussions on the oracle transfer learning procedure depend heavily on the representational similarity conditions, i.e., \eqref{eq:rep-sim1} and \eqref{eq:rep-sim2}. Otherwise, knowledge transfer estimators might have a much poorer performance than single-task learning using the target dataset alone. This is known as \emph{negative transfer} in the literature \citep{torrey2010transfer}. In this section, our aim is to provide solutions to avoid negative representational transfer. It naturally falls into two sub-problems, as \eqref{eq:rep-sim1} and \eqref{eq:rep-sim2} can be violated in the following two different ways.

 First, for some shared representations $U$ and $V$, there could exist some $\Theta_k^* = U_k^*\Lambda_k^*(V_k^*)^{\top}$ such that $\|U_{\perp}^{\top}U^*_k\|_F\gg h$ or/and $\|V_{\perp}^{\top}V^*_k\|_F\gg h$. Hence, blindly including all source datasets to estimate shared representations might cause problems. Thus, the first task is to acquire a reliable estimate of the shared representations given potentially harmful source datasets. In principle, there shall be informative sources for $U$, denoted as $\cI_U\subseteq[K]$ and informative sources for $V$, denoted as $\cI_V\subseteq[K]$, and $\cI_U$ does not necessarily equal to $\cI_V$. Therefore, in the rest of this section, we separately discuss $\cI_U$ and $\cI_V$, instead of a joint informative set $\cI$ as in the former sections.

Second, in \eqref{eq:rep-sim1} and \eqref{eq:rep-sim2}, it is assumed that the column and row subspaces of the target matrix $\Theta_0^*$ are in $\span(U)$ and $\span(V)$, respectively. However, if this is not the case and, say, $\span(U_0^*)$ is actually far away from $\span(U)$, then the target estimation step in Algorithm \ref{alg:one-round} would probably result in negative transfer. Hence, we also aim to provide a positive transfer warranty for the target task even if the shared representations are of no benefit to the target study.

In the following assume that we already acquire the unbiased matrix-valued dataset $\cD_{[K]}:=\{\tilde{\Theta}_k,k\in[K]\}$ and their corresponding left and right singular subspaces, denoted as $\tilde{P}_k$ and $\tilde{Q}_k$, respectively. We address these two sub-problems mentioned above accordingly as follows.

\subsection{Selective subspace integration}
For the first problem, we reliably retrieve the shared representations by simultaneously selecting those informative sources. We focus on $\span(U)$ here because of symmetry. For $k\in \cI_U$, the similarity condition $\|U_{\perp}^{\top}U^*_k\|_F\leq h$ is equivalent to 
\begin{equation}\label{eq:dk}
    d_k:= r_k-\tr(P_k^*P_{U}),\quad \text{such that}\quad 0\leq d_k\lesssim h^2,
\end{equation}
where $P_k^*=U_k^*(U_k^*)^{\top}$. That is, if $\span(U_k^*)$ is sufficiently close to $\span(U)$, then $d_k$ is small, implying a larger $\tr(P_k^*P_{U})$ from \eqref{eq:dk}. Meanwhile, for the oracle case such that $\cI_U$ is known, we acquire the shared linear representations $P_{\hat{U}}=\hat{U}\hat{U}^{\top}$ from \eqref{eq:GB-os}, and
\begin{equation}\label{eq:oraGB2_U}
P_{\hat{U}}=\argmax_{P\in\cG(p,p_0)}\tr(\tilde{\Sigma}_UP)=\argmax_{P\in\cG(p,p_0)}\frac{1}{N_U}\sum_{k\in\cI_U}n_k\tr(\tilde{P}_kP),
\end{equation}
where $N_U:=\sum_{k\in\cI_U}n_k$ and $\cG(p,p_0)$ is defined as the set of $p_0$-dimensional linear subspaces of $\reals^p$. Inspired by both \eqref{eq:dk} and \eqref{eq:oraGB2_U}, we follow \cite{li2024knowledge} and focus on the following rectified problem that is capable of selecting useful datasets and estimating the shared representation simultaneously:
\begin{equation}\label{eq:GBkmeans_U}
P_{\hat{U}_{\tau}}=\argmax_{P\in \G(p,p_0)}\frac{1}{N_{[K]}}  \sum_{k\in[K]}n_k\max\left\{\tr(\tilde{P}_kP),r_k-\tau_U\right\},
\end{equation}
for $\tau_U\in [0,p_0]$, $P_{\hat{U}_{\tau}}:=\hat{U}_{\tau}\hat{U}_{\tau}^{\top}$ and $N_{[K]}=\sum_{k\in[K]}n_k$. Clearly, if $\tau_U=0$, then \eqref{eq:GBkmeans_U} is an optimization over a constant and no source information is included. Meanwhile, if $\tau_U=p_0$, then \eqref{eq:GBkmeans_U} degenerates to the Grassmannian barycenter method such that all sources are included blindly. Ideally, we expect some suitable $\tau_U\in (0,p_0)$ that can separate these informative sources from harmful ones. 

To numerically solve \eqref{eq:GBkmeans_U}, we resort to the rectified Grassmannian K-means procedure from \cite{li2024knowledge}. Given any initialization $P_0\in \cG(p,p_0)$, in the $t$-th iteration step we have $P_{t-1}$ from the previous step for $t\geq 1$. Then, we select the informative sources according to $\cI^{t}_U=\{k\in[K]\mid \tr[\tilde{P}_kP_{t-1}]\geq r_k-\tau_U \}$. We perform the Grassmannian barycenter method for $k\in \cI^{t}_U$, updating $P_{t-1}$ to $P_t$. We proceed with the iteration until we arrive at convergence. Similarly, for $\span(V)$, we could acquire $P_{\hat{V}_\tau}:=\hat{V}_\tau\hat{V}_\tau^{\top}$ given some $\tau_V\in [0,q_0]$.

We give a brief remark on the dataset selection capability of the rectified problem \eqref{eq:GBkmeans_U}. We show that the oracle estimator from \eqref{eq:oraGB2_U} is also a local maximum of \eqref{eq:GBkmeans_U} with high probability under certain separability conditions. We only discuss on $\span(U)$ here as that on $\span(V)$ is basically identical.

\begin{assumption}[Separable non-informative sources]\label{assum:5}
For $d_k= r_k-\tr(P_k^*P_{U})$ with respect to $k\in \cI_U^c:=[K]\setminus \cI_U$, assume that $d_k\geq d_\tau$ for some $d_\tau>0$.
\end{assumption}

\begin{corollary}[Local maximum]\label{theo:nora}
    Under the assumptions of Theorem \ref{theorem:main} (replacing $\cI$ by $\cI_U$ whenever needed) and Assumption \ref{assum:5}, if $h+(rp\log p/N_{U})^{1/2}+\max_{k\in \cI_U^c}\{(rp\log p/n_k)^{1/2}\}=o(d_\tau)$ as $n_k, p,q\rightarrow \infty$, choose any $\tau_U\in [(1-c)d_\tau, cd_\tau]$ for some $c\in (0.5,1)$, then $P_{\hat{U}}$ from \eqref{eq:oraGB2_U}  is a local maximum of \eqref{eq:GBkmeans_U} with probability tending to 1.
\end{corollary}

\subsection{Optional knowledge transfer}
For the second problem, the new-arrived target matrix may not be well represented by the jointly estimated $\hat{U}_{\tau}$ and $\hat{V}_{\tau}$. Thus, further emphasis lies on whether $\hat{U}_{\tau}$  and $\hat{V}_{\tau}$ can be utilized to enhance the performance of the target task. Motivated by \eqref{eq:dk} and the corresponding rectified problem above, we first acquire a matrix completion estimator $\dot{\Theta}_0$ using only the target dataset. The leading $r_0$-dimensional left and right singular subspaces of $\dot{\Theta}_0$ are defined as $\span(\dot{U}_0)$ and $\span(\dot{V}_0)$, respectively. By comparing the similarity of $\span(\dot{U}_0)$ and $\span(\dot{V}_0)$ with $\span(\hat{U}_\tau)$ and $\span(\hat{V}_{\tau})$ respectively before the target estimation step, we judge whether transferring the representational knowledge to the target matrix is helpful. Specifically, let $P_{\dot{U}_0}:=\dot{U}_0\dot{U}_0^{\top}$ and $P_{\dot{V}_0}:=\dot{V}_0\dot{V}_0^{\top}$, for $\delta_U\in [0,r_0]$ and $\delta_V\in [0,r_0]$, let
\begin{equation}\label{U: transfer robust}
\hat{U}_\tau^{\delta}=
\left\{\begin{array}{l}
\hat{U}_{\tau},\quad \text{if } \tr(P_{\dot{U}_0}P_{\hat{U}_{\tau}})\geq r_0-\delta_U,\\
\dot{U}_0,\quad\  \text{if } \tr(P_{\dot{U}_0}P_{\hat{U}_{\tau}})< r_0-\delta_U,
\end{array}\right.\quad \hat{V}_\tau^{\delta}=
\left\{\begin{array}{l}
\hat{V}_{\tau},\quad \text{if } \tr(P_{\dot{V}_0}P_{\hat{V}_{\tau}})\geq r_0-\delta_V,\\
\dot{V}_0,\quad\  \text{if } \tr(P_{\dot{V}_0}P_{\hat{V}_{\tau}})< r_0-\delta_V.
\end{array}\right.
\end{equation}

\begin{algorithm}
\caption{The pseudo algorithm for non-oracle representational transfer.}\label{alg:npse}
\begin{algorithmic}[1]
\REQUIRE ~~\\
    $(X_{k,i}, y_{k,i})$ for $k\in\cI=[K]$; cut-off dimensions $p_0$, $q_0$; tuning parameters $\tau_U$, $\tau_V$, $\delta_U$, $\delta_V$;\\
\ENSURE ~~\\
    \STATE 1. unbiased dataset: for $k\in\cI$, acquire an unbiased estimator $\tilde{\Theta}_k$ of $\Theta_k^*$, construct the unbiased matrix-valued dataset $\cD_{\cI} :=\{\tilde{\Theta}_k, k\in \cI\}$;\\
    \STATE 2. selective and optional transfer: acquire the selective and optional linear representations $\hat{U}_\tau^{\delta}$ and $\hat{V}_\tau^{\delta}$ following the procedures of this section;\\
    \STATE 3. target estimation: for $\hat{Z}_{0,i}^{\tau,\delta}:=(\hat{U}_\tau^{\delta})^{\top}X_{0,i}\hat{V}_\tau^{\delta}$, apply ordinary least squares to the projected dataset $\{(\hat{Z}^{\tau,\delta}_{0,i},y_{0,i})\}_{i=1}^{n_0}$, so as to obtain the $p_0\times q_0$ parameter matrix $\hat{\Gamma}_0^{\tau,\delta}$;\\
\RETURN $\hat{\Theta}_{0}^{\tau,\delta}= \hat{U}_\tau^{\delta}\hat{\Gamma}_0^{\tau,\delta}(\hat{V}_\tau^{\delta})^{\top}$.
\end{algorithmic}
\end{algorithm}
Finally, we plug in $\hat{U}_\tau^{\delta}$ and $\hat{V}_\tau^{\delta}$ into the target estimation step. The Non-ORAcle (Nora) representational transfer procedures are summarized in the pseudo Algorithm \ref{alg:npse}. We end this section with a few remarks. First, if $P_{\dot{U}_0}$ is estimated using our default matrix completion method \citep{xia2021statistical}, it shall be mathematically equivalent to the estimator acquired by plugging both $\dot{U}_0$ and $\dot{V}_0$ into the target estimation step by construction. That is to say, if we choose the most conservative $\delta_U=\delta_V =0$ so that $\hat{U}_\tau^{\delta}=\dot{U}_0$ and $\hat{V}_\tau^{\delta}=\dot{V}_0$, the resulting estimator from Algorithm \ref{alg:npse} is equivalent to the target matrix completion estimator. Meanwhile, analogously to the claims in Corollary \ref{theo:nora}, if $h+(rp\log p/N_{U})^{1/2}+(rp\log p/n_0)^{1/2}=o(\delta_U)$ for some $\delta_U>0$, the probability that we correctly identify and include the oracle $P_{\hat{U}}$ into target estimation tends to $1$ as $n_0, n_k, p,q\rightarrow \infty$. Almost identical arguments hold for $\span(V)$. In short, Algorithm \ref{alg:npse} is capable of simultaneously selecting useful sources and transferring representational information with positive warranty.

\section{Numerical Simulation}\label{sec:method_comparison}

In this section, we perform numerical simulation to validate the usefulness of the proposed methods. We generate the datasets as follows. We first take a $p_0$ dimensional $\span(U)$ uniformly, that is, with respect to the Haar measure, from $\cG(p,p_0)$. Similarly, we uniformly sample a $q_0$ dimensional $\span(V)$ from $\cG(q,q_0)$. As any subspace can be uniquely represented by the orthogonal projector from $\reals^p$ onto itself, we write the corresponding projection matrices as $P_U$ and $P_V$. 

Then, define $\text{GOE}(p)$ as the $p\times p$ symmetric random matrix with off-diagonal entries taken independently from $\mathrm{N}(0,(2p)^{-1})$ and the diagonal entries taken independently from $\mathrm{N}(0,(p)^{-1})$, for $\Theta_k^* = U_k^*\Lambda_k^*(V_k^*)^{\top}$, $k\in \{0\}\cup [K]$, we generate $U_k^*$ and $V_k^*$ in the following four ways: ({\romannumeral1}) $U_k^*$ and $V_k^*$ are acquired by taking the leading $r_k$ eigenvectors of $P_U+hG_k$ and $P_V+hH_k$ respectively, where $G_k$ is generated independently from $\text{GOE}(p)$, while $H_k$ is generated independently from $\text{GOE}(q)$; ({\romannumeral2}) $U_k^*$ is the leading $r_k$ eigenvectors of $P_U+hG_k$, while $V_k^*$ is sampled uniformly from $\cG(q,r_k)$; ({\romannumeral3}) $U_k^*$ is sampled uniformly from $\cG(p,r_k)$, while $V_k^*$ is the leading $r_k$ eigenvectors of $P_V+hH_k$; ({\romannumeral4}) $U_k^*$ and $V_k^*$ are sampled uniformly from $\cG(p,r_k)$ and $\cG(q,r_k)$, respectively. For all $4$ cases, $\Lambda_k^*$ is generated as a $r_k\times r_k$ diagonal matrix whose diagonal elements are drawn independently from $\text{Uniform}(1,2)$ distribution.

For the sources, that is, for $k\in [K]$, we generate $K/2$ of $\Theta_k^*$ from ({\romannumeral1}), and $K/6$ each from ({\romannumeral2}), ({\romannumeral3}) and ({\romannumeral4}), so that not all left and right singular subspaces from the sources are informative. Meanwhile, we take the target matrix $\Theta_0^*$ from ({\romannumeral1}), ({\romannumeral2}) and ({\romannumeral4}), leading to three scenarios: scenario A where $\Theta_0^*$ is taken from ({\romannumeral1}), and it can take advantage of both column and row representations from the informative sources; scenario B where $\Theta_0^*$ is taken from ({\romannumeral2}), and only one side of the representational information is useful; scenario C where $\Theta_0^*$ is taken from ({\romannumeral4}), and the target completion task cannot benefit from the sources and we shall be cautious about the negative transfer. We set $p=q=100$, $p_0=q_0=5$, $r_k=3$ and $h=0.1$. For $k\in\{0\}\cup [K]$ where $k\in \{24,36,48,60,72\}$, we uniformly sample $n_k=2500$ entries of $\Theta_k^*$ contaminated by additive noise with $\sigma_k = 1$.

\begin{figure}
	\centering

    \begin{minipage}{0.3\linewidth}
		\centering
	\includegraphics[width=0.9\linewidth]{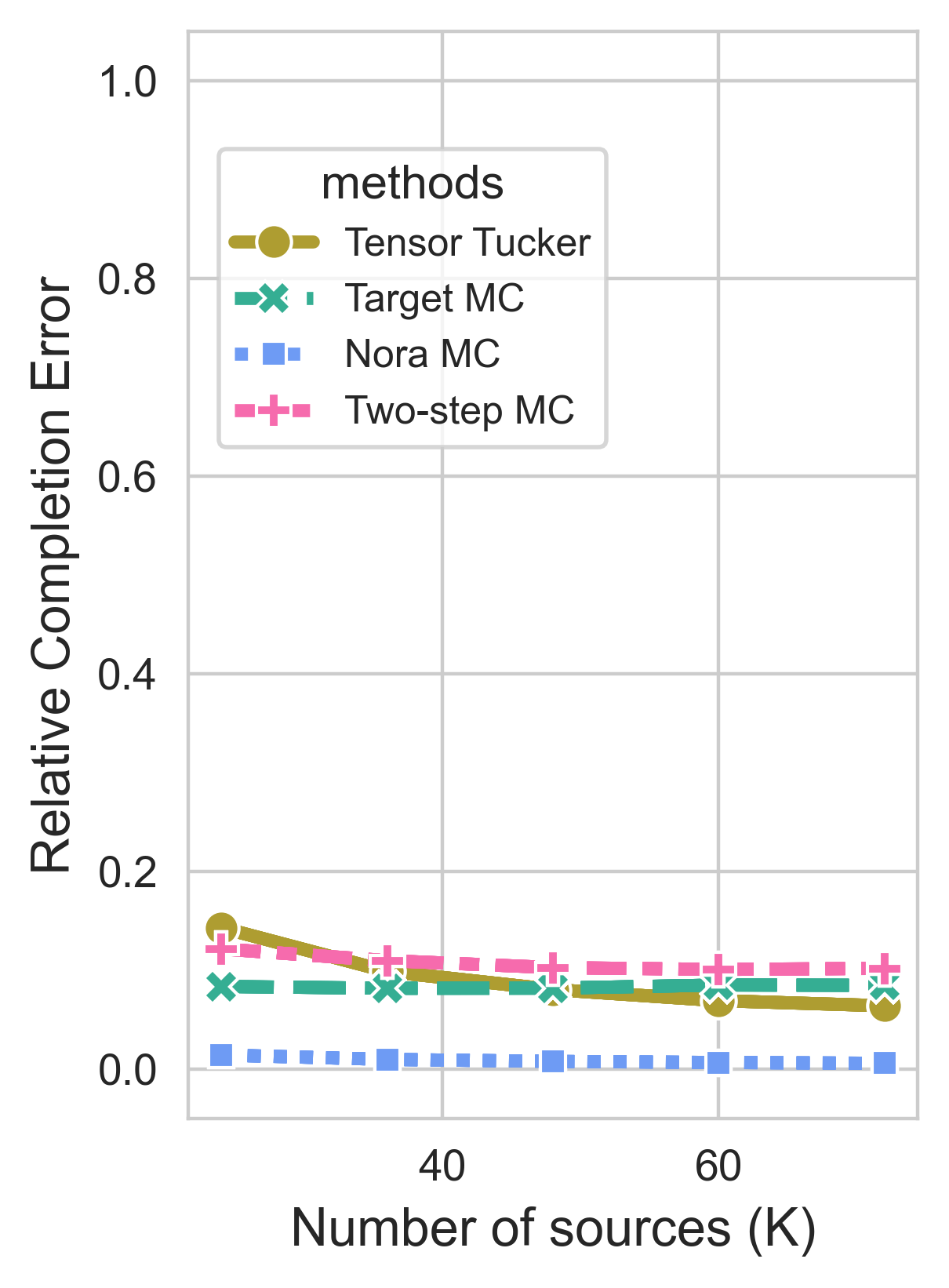}
	\end{minipage}
        \begin{minipage}{0.3\linewidth}
		\centering
	\includegraphics[width=0.9\linewidth]{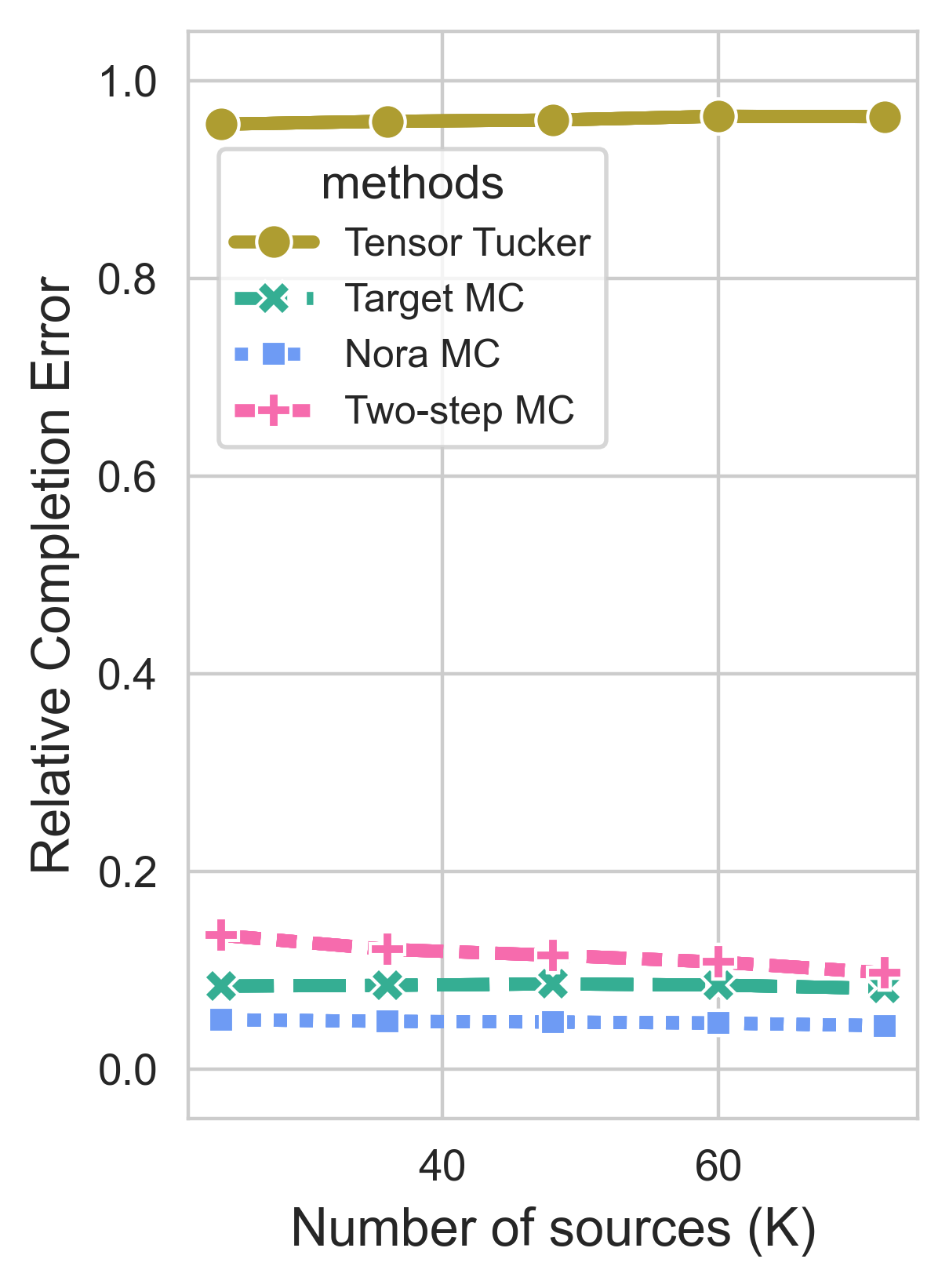}
	\end{minipage}
	\begin{minipage}{0.3\linewidth}
		\centering
	\includegraphics[width=0.9\linewidth]{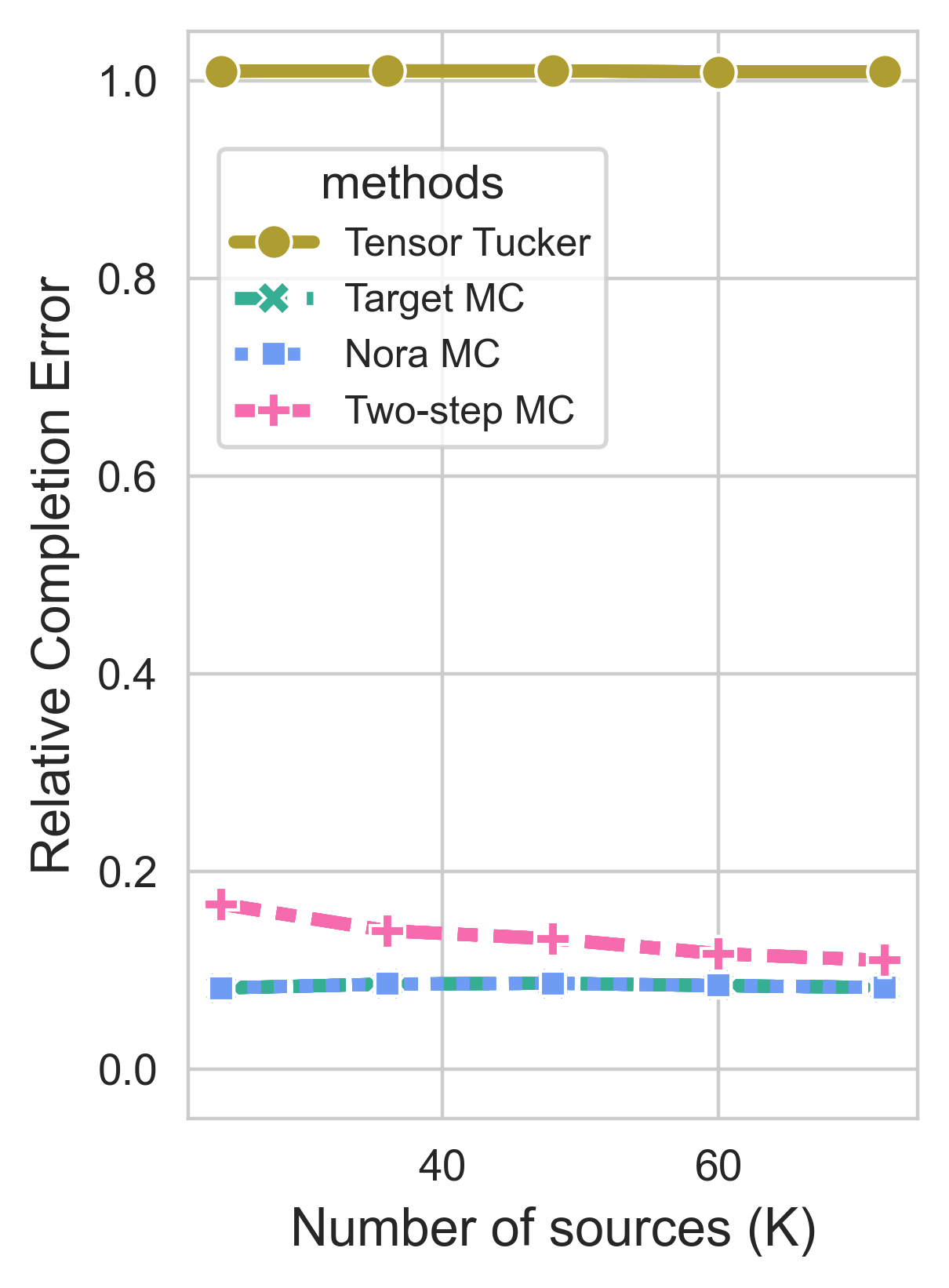}
	\end{minipage}
\caption{Comparison of various methods under different scenarios, i.e., scenario A, B and C from left to right. The proposed non-oracle knowledge transfer algorithm demonstrates its advantage across the three given settings. \label{fig-sim-main} }
\end{figure}

We compare Algorithm \ref{alg:npse} with various methods, where the unbiased matrix-valued dataset $\cD_{[K]}$ is generated by the default $J$-fold sample splitting with $J=5$. For the competitors, we consider the following. 1. The target matrix completion estimator by applying our default matrix completion method, i.e. the calibrated Grassmannian gradient descent algorithm without sample splitting from \cite{xia2021statistical}, on the target dataset; 2. the two-step matrix completion estimator by first using the default matrix completion method on the pooled observed samples and then using the same default matrix completion to estimate the contrast by the target dataset; 3. the first slice of the tensor-Tucker completion estimator using the power iteration method from \cite{xia2021statistically}, treating all observations as if they are from a $(K+1)\times p\times q$ stacked tensor with missing entries.

In Figure \ref{fig-sim-main}, we report the average relative completion error of each estimator of the target matrix, defined as $\|\cdot-\Theta_0^*\|_F^2/\|\Theta_0^*\|_F^2$, based on $100$ replications. We can see that Algorithm \ref{alg:npse} shows an advantage over competing methods in all settings. In particular, in both scenario B and scenario C, our selective and optional transfer algorithm performs not worse than the target matrix completion estimator, demonstrating its robustness against useless source information, while the blind tensor-Tucker completion method leads to devastating negative transfer.

\section{Real Data Experiment}

In this section, we provide empirical evidence on the practical usefulness of the proposed method. To do this, we first take the COVID-19 computed tomography (CT) chest scans dataset from \cite{zhang2022low}, which includes 349 gray-scale chest scan images from COVID-19 positive patients. Typically, a gray-scale image is stored as a matrix. The elements of the matrix depict the color depth of the pixels, ranging from $0$ to $255$. We first resize all the images to the size of $150\times 150$ using the Python package \href{https://pillow.readthedocs.io/en/stable/}{\texttt{PIL}}. Then we standardize the pixel values by subtracting all the elements by $255/2$ and then dividing them by $255$.

\begin{figure}[h]
  \centering
    \begin{minipage}[t]{0.985\linewidth}
  \centering
  \includegraphics[width=4.5in]{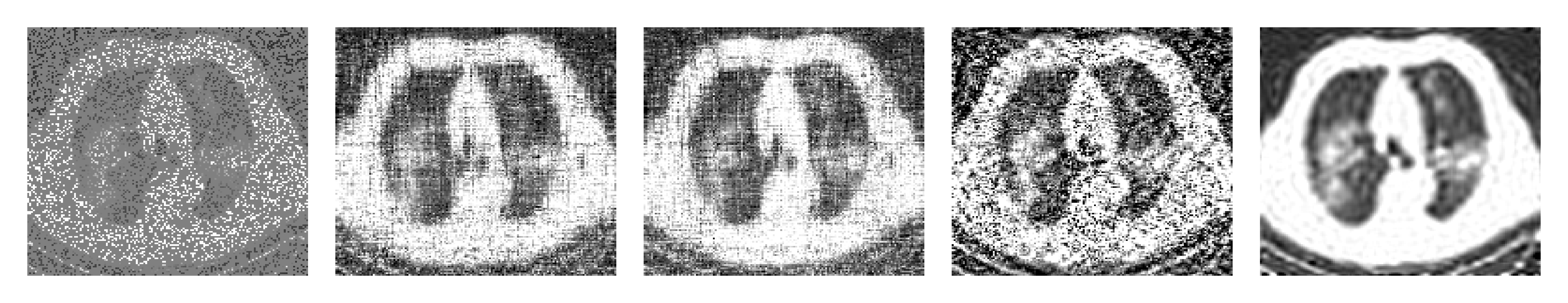}
  \includegraphics[width=4.5in]{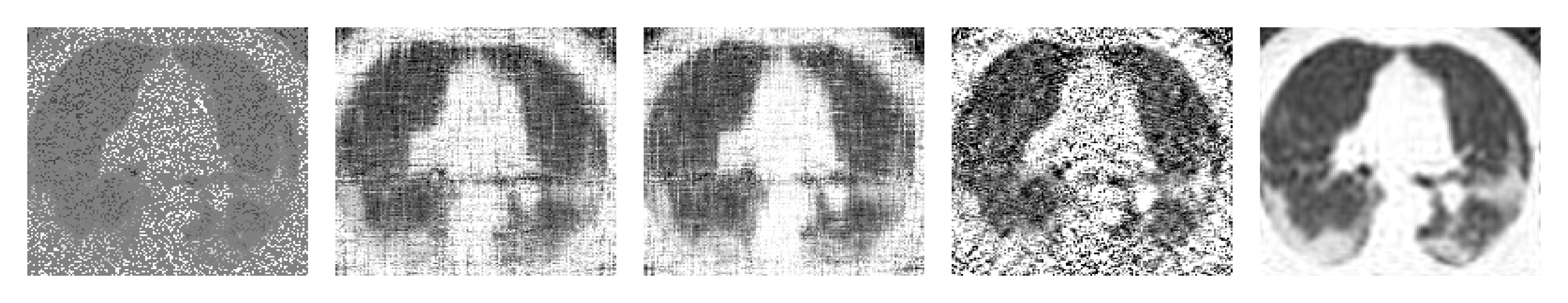}

  \includegraphics[width=4.5in]{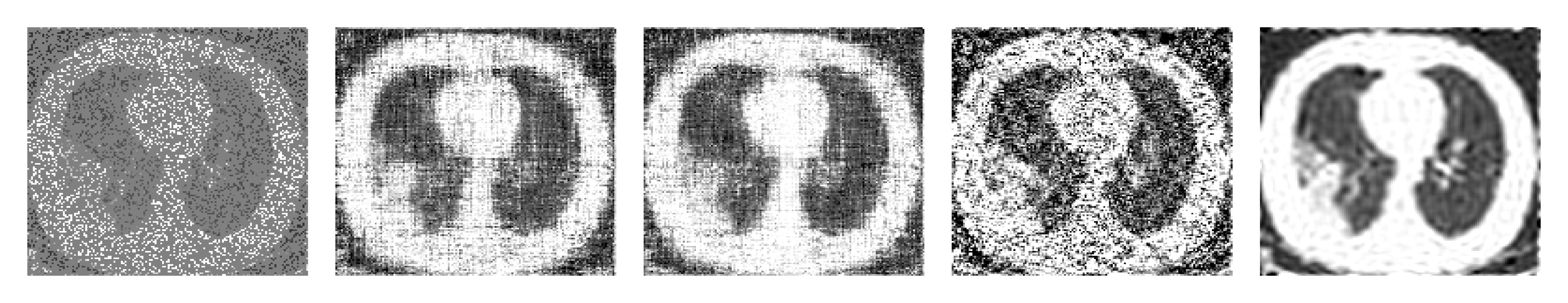}
  \end{minipage}
  \caption{For each panel, from left to right are respectively: (a) the partially observed image; (b) the target completed image using the target dataset only; (c) the two-step completed image by debiasing the Euclidean mean of the sources; (d) the first slice of the tensor-Tucker completion estimator and (e) the representational transferred completed image by utilizing the subspace information from the sources.}\label{fig:covid-completion}
\end{figure}

For visual illustration, each time we choose one of the three typical CT images to be the target, denoted by $\Theta_0^*$. We uniformly sample $n_0=6000$ elements from $\Theta_0^*$ with $\sigma_0=1$, forming the target dataset $\{(X_{0,i},y_{0,i})\}_{i=1}^{n_0}$. For each target dataset, we randomly take $K = 30$ fully observed sources from the dataset, denoted as $\{\Theta_k^*\}_{k=1}^{30}$. We plot the following images in each panel of Figure \ref{fig:covid-completion}: (a) the partially observed target image; (b) the target completed image using the target dataset only; (c) the two-step completed image by debiasing the Euclidean mean of the sources, namely $\sum_{k=1}^{K}$ $\Theta_k^*/K$; (d) the tensor-Tucker completed image by taking the first slice of the $(K+1)\times p\times q$ stacked tensor; and (e) the representational transferred completed image  integrating the subspace information from the sources by Algorithm \ref{alg:one-round}. To acquire the left and right subspaces, namely $\hat{U}$ and $\hat{V}$, from the sources, we apply the one-round Grassmannian barycenter method with $p_0=q_0=40$ and $r_k = 20$. The cut-off dimensions are determined in a fully unsupervised manner following \cite{2dgb}. We can see that the target completed image, the two-step completed image, and the tensor-Tucker completed image are all blurry and difficult to interpret. In contrast, the completed image after representational transfer is able to retrieve most of the pathological features of the original image. Figure \ref{fig:covid-completion} visually justifies the representational similarity in this analysis of the real dataset.

\section{Discussion}
For saving space, we delegate extensional arguments on the fully observed sources and the determination of cut-off dimensions to the supplementary material. We also report additional simulation results that validate the statistical error rate in Theorem \ref{theorem:main} and the asymptotic normality of the post-transfer bilinear forms estimators. A real data example on functional magnetic resonance imaging (fMRI) of patients with attention deficit hyperactivity disorder (ADHD) at different sites is also provided. All the proofs of our theoretical results can be found in the supplementary material.
For further extensions, it is rather straightforward to extend the procedures in this work to higher-order tensors. Meanwhile, it will be intriguing to take advantage of the kernel methods to integrate the non-linear representational information in an unsupervised manner \citep{couillet2022random,he2024representation}.

\bibliographystyle{apalike}
\bibliography{ref}

\end{document}